\documentclass{article}
\usepackage{fullpage}

\usepackage{natbib}
\usepackage{subfig}

\usepackage[utf8]{inputenc} %
\usepackage[T1]{fontenc}    %
\usepackage{url}            %
\usepackage{booktabs}       %
\usepackage{amsfonts}       %
\usepackage{nicefrac}       %
\usepackage{microtype}      %
\usepackage{xcolor}         %
\usepackage{authblk}
\usepackage{amsmath}
\usepackage{amssymb}
\usepackage{mathtools}
\usepackage{amsthm}
\usepackage{hyperref}
\hypersetup{colorlinks,allcolors=black}

\theoremstyle{plain}
\newtheorem{theorem}{Theorem}[section]

\newtheorem{lemma}[theorem]{Lemma}
\newtheorem{corollary}[theorem]{Corollary}
\theoremstyle{definition}
\newtheorem{definition}[theorem]{Definition} 
\newtheorem{assumption}[theorem]{Assumption}
\theoremstyle{remark}
\newtheorem{remark}[theorem]{Remark}

\newcommand{\R}{\mathbb{R}}

\newcommand{\E}{\mathbb{E}}

\newcommand{\Var}{\mathrm{Var}}
\newcommand{\script}[1]{\mathcal{#1}}

\title{Offline Policy Evaluation for Reinforcement Learning with Adaptively Collected Data}

\author[1]{Sunil Madhow}
\author[1]{Dan Qiao}
\author[1]{Ming Yin}
\author[1]{Yu-Xiang Wang}
\affil[1]{UC Santa Barbara}
\date{}

\begin{document}

\maketitle

\begin{abstract}
  Developing theoretical guarantees on the sample complexity of offline RL methods is an important step towards making data-hungry RL algorithms practically viable. Currently, most results hinge on unrealistic assumptions about the data distribution -- namely that it comprises a set of i.i.d. trajectories collected by a single logging policy. We consider a more general setting where the dataset may have been gathered adaptively. We develop theory for the TMIS Offline Policy Evaluation (OPE) estimator in this generalized setting for tabular MDPs, deriving high-probability, instance-dependent bounds on its estimation error. We also recover minimax-optimal offline learning in the adaptive setting. Finally, we conduct simulations to empirically analyze the behavior of these estimators under adaptive and non-adaptive regimes.
\end{abstract}

\section{Introduction} \label{introduction}

Offline Reinforcement Learning (RL), which seeks to perform standard RL tasks using a pre-existing dataset of interactions with an MDP, is a key frontier in the effort to make RL methods more widely applicable. The ability to incorporate existing data into RL algorithms is crucial in many promising application domains. In safety-critical areas, such as autonomous driving \citep{kivan_drive}, the randomized exploration that characterizes online algorithms is not ethically tolerable. Even in lower-stakes applications, such as advertising \citep{cai2017real}, naively adopting online algorithms could mean throwing away vast reserves of previously-collected data. The development of efficient offline algorithms promises to broaden RL's applicability by allowing practitioners to exercise some much needed domain-specific control over the training process.

Given a dataset, $\script D$, of interactions with an MDP $\mathcal{M}$, two tasks that we may hope to achieve in offline RL are Offline Policy Evaluation \citep{yin20-0} and Offline Learning \citep{lange2012batch}. In Offline Policy Evaluation (OPE), we seek to estimate the value of a target policy $\pi$ under $\script M$. In Offline Learning (OL), the goal is to use $\script D$ to find a good policy $\pi \in \Pi$ where $\Pi$ is some policy class. In this paper, we largely focus on OPE.

The question of how and when it is possible to perform OPE and OL given a specific dataset is the subject of a flourishing research movement \citep{lange2012batch, Raghu19matters, Le19_constraints, Xie20batch, duan20, yin20-0, yin20-1, jin2021pessimism, Jin22Overlap, ,qiao2022offline,zhang2022off}. In order for $\script D$ to be a rich enough dataset to learn from, it must explore the MDP well. Excellent results have been derived assuming that $\script D$ consists of i.i.d. trajectories distributed according to some logging policy $\mu$, where $\mu$ has good exploratory properties. However, it is difficult to justify the imposition of these assumptions on our data. Empirically, it is clear that the generation of a usable $\script D$ is a significant bottleneck in offline RL. In practice, the gathering of useful datasets is best done by running adaptive exploration algorithms (see, for example, \cite{Lambert22Challenges}'s use of ``curiosity''). In the case of self-driving cars, \emph{any} dataset will be rife with adaptivity due to human interventions and the need to iterate policies. If the theory of offline RL is to be relevant in real-world use-cases, it must extend to adaptively collected datasets. 

Our object in this paper is to develop a systematic understanding of the role of adaptivity in offline RL. In Section \ref{related work}, we discuss the problems posed by the i.i.d. assumption from a more rigorous perspective. We proceed to introduce a more general setting for OPE, Adaptive OPE (AOPE), where we allow each trajectory to be distributed according to a different logging policy, which may depend on previous data. 

In addition to the motivating examples given above, here are some scenarios that AOPE covers but OPE does not:
\begin{enumerate}
    \item The dataset $\script D$ was gathered by humans, and therefore influenced by a number of unobserved factors and historical data. For example, a doctor prescribing medicine may make a determination based on her conversation with the patient and prior experience \citep{Yu_med}.
    \item The dataset $\script D$ has been gathered by a reward-free exploration algorithm \citep{jin20rewardfree, wang2020reward, qiao2022near}. This dataset will have excellent exploratory properties, but is very intradependent.
\end{enumerate}

In Section \ref{motivation}, we explain how \cite{yin20-1}'s minimax-optimal results for both Uniform OPE and efficient learning lead directly to optimal results for the adaptive case. 

In Section \ref{theory_sec}, we proceed to derive more evocative \emph{instance dependent} bounds on the estimation error $|\hat{v}^\pi - v^\pi|$. By introducing a novel technique for performing concentration in the face of adaptivity, we prove that, with high-probability, $|\hat v^\pi - v^\pi|$ is governed by a dominant term of \[\widetilde{O}\left(\sum_{h = 1}^H \sum_{s, a} d^\pi_h(s, a)\sqrt{\frac{\Var_{s' \sim P_{h + 1}(\cdot | s, a)}[V^\pi_{h + 1}(s')]}{n_{h, s, a}}}\right),\]
where $\widetilde{O}(\cdot)$ absorbs logarithmic terms and constants, $n_{h,s,a}$ is the visitation number of $(s,a)$ at time step $h$.

We note that our instance-dependent bounds may significantly outperform the minimax-optimal bound on certain instances, but they fail to recover optimal \emph{worst-case} rates.

In Section \ref{simulations}, we empirically study how adaptivity can improve or degrade the performance of our estimator. 

\subsection{Related Work}\label{related work}
To the best of our knowledge, we are the first to consider OPE for reinforcement learning under adaptive data. However, in the study of bandits and RL, Off-Policy Evaluation has been an area of interest for more than a decade \citep{dudik2011doubly, jiang2016doubly, wang2017optimal, thomas2016dataefficient, yin20-0, yin20-1}. In RL, existing work adopts the setting where $\script D = \{\tau_i \sim \mu\}$ is a collection of i.i.d. trajectories. In this setting, bounds on the performance of OPE or OL algorithms are given in terms of an exploration parameter, like:

\begin{equation}\label{sing-pol-dm}d_m = \min_{h, s, a: d^\pi_h(s, a) > 0}d^\mu_h(s, a). \end{equation} Estimators that closely match lower bounds on relevant metrics have been established in existing work \citep{duan20, yin20-1, Xie20batch, Le19_constraints}.

The multi-logger setting, where $\script D = \{\tau_i \sim \mu^i\}$ for $\mu^i$ statically chosen, is a straightforward generalization of the single-logger setting, and so we refer to this problem as Non-Adaptive OPE (NOPE).  

However, the practicability of such bounds has been challenged. \cite{xiao21} point out that it is difficult in practice to find a logging policy with a reasonable exploration parameter. In what they consider a more realistic, ``tabula rasa" case (where the logging policy is chosen without knowledge of the MDP), they show a sample complexity exponential in $H$ and $S$ to be necessary in offline learning. Current results also fail to address the motivating application of learning from existing, human-generated data, which we would not expect to be identically distributed, or even independent (as the data collected in trajectory $j$ almost certainly influences future policies $\mu^{j+1}, ...$). Results for OPE with adaptive data would address both of these concerns.

While we do not know of any work that studies OPE with adaptively collected data, \cite{jin2021pessimism} study the problem of Offline Learning with Pessimistic Value Iteration under adaptive data for linear MDPs. Their results do not cover OPE, and are loose when specialized to tabular MDPs. \cite{Jin22Overlap} cover the problem of learning from adaptive data for contextual bandits. The generalization of such results to reinforcement learning is highly nontrivial, and the approach we take to the problem is largely unrelated. For multiarmed bandits, \cite{Shin19Bias} study how adaptive exploration schemes like optimism can lead to bias in estimated arm values. In their work, they imagine a data-collection model whereby a table is populated with data before any experiments begin. We make use of a similar model, generalized to the RL setting. This work also provides inspiration for our numerical simulations.

\begin{figure*}
\centering
  \includegraphics[width=0.8\linewidth]{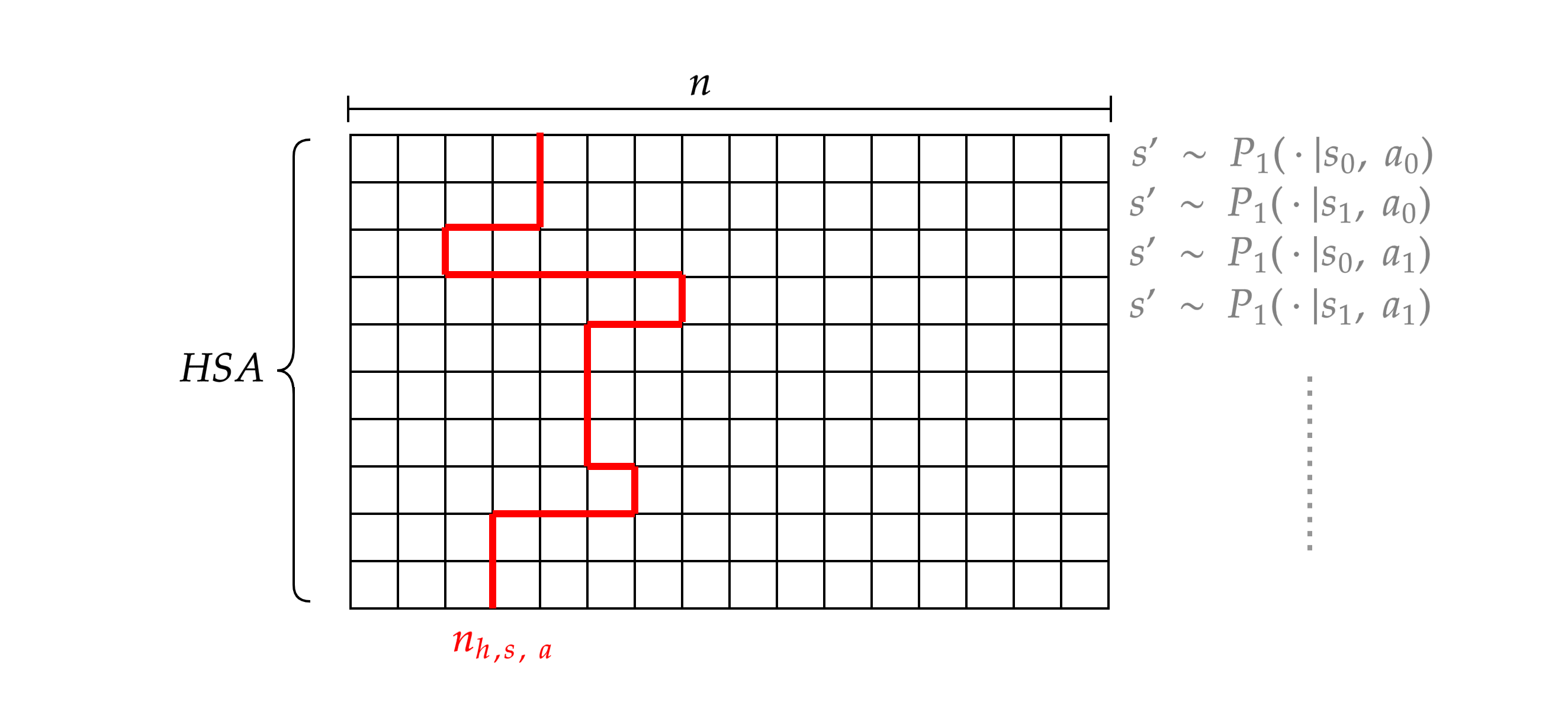}
    \caption{An illustration of the tape view of adaptive data collection. Each row $(h, s, a)$ should be thought to contain $n$ i.i.d. samples from $P_{h + 1}(\cdot | s, a)$. The red ``frontier'' tracks how many samples have, in fact, been used by the logger (this quantity is always bounded by $n$). }
    \label{tape pic}
\end{figure*}

\subsection{Novel Contributions}
We now highlight what in our view are the key contributions of this paper.

In the proof of our high-probability bounds on estimation error, $|\hat{v}^\pi - v^\pi|$, we utilize a novel proof technique for dealing with intradependent datasets. Leveraging the fact that influence of the logger's adaptivity on the dataset is limited to the quantities $\{n_{h, s, a}\}$, we  consider an equivalent data-collection model, defined by a machine with $H \times S \times A$ rolls of tape. Each tape contains $n$ i.i.d. samples from the distribution $P_h(\cdot | s, a)$. When the logger visits $(h, s, a)$, a corresponding $s'$ is read off the tape, $(h, s, a, s')$ is added to the dataset, and the tape is advanced. This model makes no distinction between trajectories, suppressing their confounding influence while retaining the problem-structure crucial to our estimator. See Figure \ref{tape pic} for an illustration.

One of our goals is to facilitate the importation of results from the non-adapive case to the adaptive case. The ``tape-machine'' model yields a prescription for selectively throwing away data in order to burn the adaptivity out of a dataset. This salvages the Martingale structure key to optimal Uniform OPE \citep{yin20-1}, and even allows one to use existing theory to perform optimal OL on the resultant dataset.

\begin{figure*}
    \centering
    \includegraphics[width=0.8\textwidth]{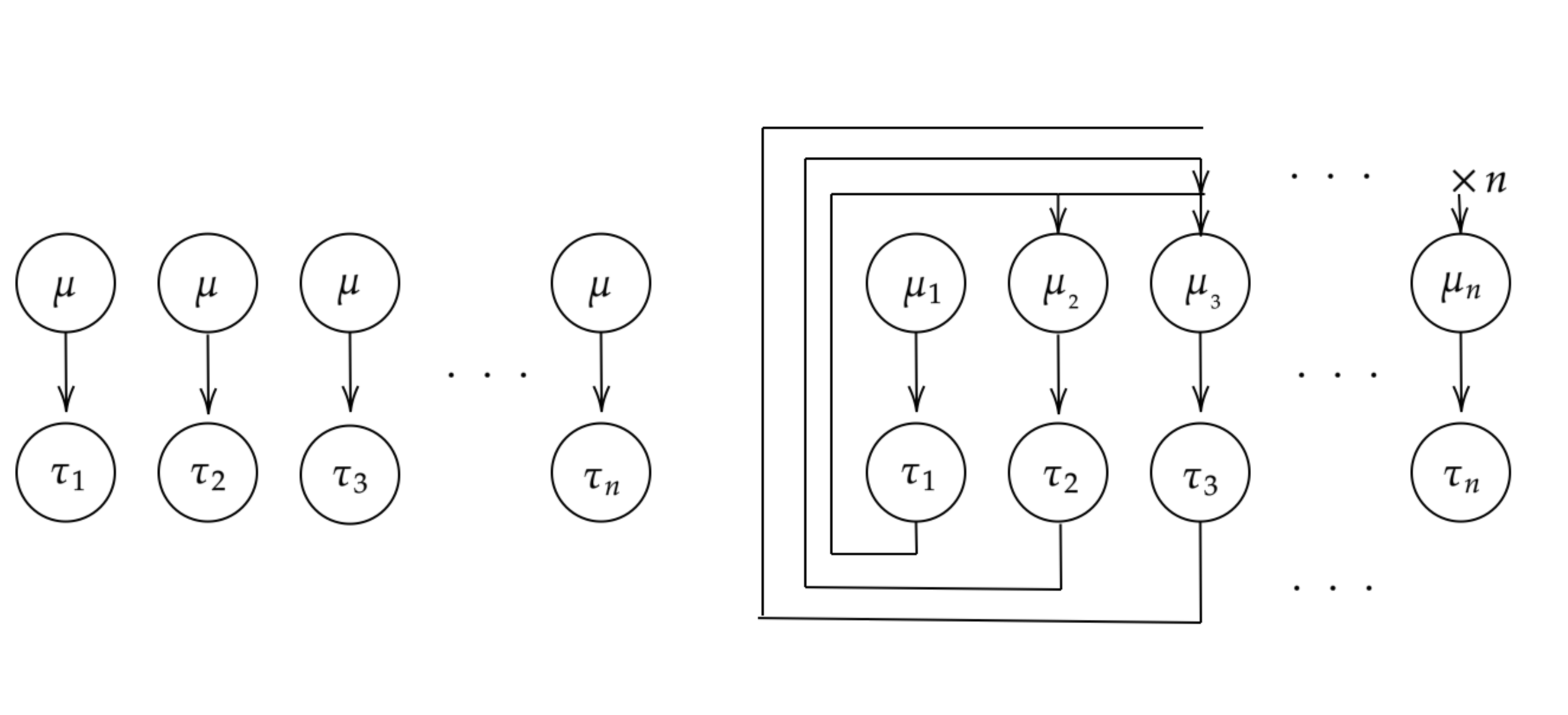}
    \caption{Non-adaptive regime (left) versus adaptive regime (right), depicted as a graphical model. We see that, in the adaptive regime, each policy depends on all previous trajectories. This induces dependence between the trajectories.}
    \label{Adaptive OPE versus OPE}
\end{figure*}

\section{Preliminaries} 

\subsection{Symbols, notation, and MDP basics. }
Let $\Delta(\script X)$ be the set of all probability distributions over $\script X$, for $|\script X| < \infty$. We denote $[H] := \{1, ... , H\}$.

A Tabular, Finite-Horizon Markov Decision Process (MDP) is a tuple $\mathcal{M}=(\script S, \script A, r, P, d_1, H)$, where $\script S$ is the discrete state space with $S:=|\script S|$, while $\script A$ is the discrete action space with $A:=|\script A|$. Its dynamics are governed by a non-stationary transition kernel, 
$P = \{P_h: \script S \times \script A \rightarrow \Delta(\script S)\}_{h = 1}^H$, where $P_h(s' | s, a)$ is the probability of transitioning to state $s' \in \script S$ after taking action $a \in \script A$ from state $s \in \script S$ at time $h \in [H]$. $r$ is a collection of reward functions $\{r_h: \script S \times \script A \rightarrow [-1, 1]\}_{h = 1}^H$. Finally, $d_1 \in \Delta(\script S)$ is the initial state distribution of the MDP and $H$ is the horizon.

A policy, $\pi$, is a collection of maps, $\{\pi_h: \script S \to \Delta(\script A)\}_{h = 1}^H$. Running a policy on an MDP will yield a trajectory $\tau_i \in (\script S \times \script A \times [-1, 1])^H$. Together, the policy and MDP induce a distribution over trajectories, as well as a Markov Chain with transitions notated as $P^\pi_h(s' | s) := \sum_a P_h(s' | s, a)\pi_h(a | s)$.

In a set of trajectories $\{\tau_i\}_{i = 1}^n$, we define $n_{h, s, a}$ to be the number of visitations to $(s, a)$ at timestep $h$ for all $h,s,a$.

$v^\pi := \E_\pi[\sum_{i = 1}^H r_i | s_1 \sim d_1]$ is the value of the policy $\pi$, where the expectation is over the $\pi$-induced distribution over trajectories. Furthermore, we define for any $\pi$ the value-function $V^\pi_h(s) := \E_\pi[\sum_{i = h}^H r_i | s_h = s]$ and Q-function $Q^\pi_h(s, a) := \E_\pi[\sum_{i = h}^H r_i | s_h = s, a_h = a]$ for $1 \leq h \leq H$.

$d^\pi_h(s, a)$ is defined to be the probability of $(s_h, a_h)$ occurring at time step $h$ in a trajectory distributed according to policy $\pi$.

\subsection{Motivation and Problem Setup}\label{motivation}
Motivated both by the negative result from \cite{xiao21} and the exigencies of real-world data, we augment our formulation of the OPE problem to more realistically accommodate intelligent choices of logging-policy. We consider this to be middle ground between the sanguine assumptions on $d_m$ common to \cite{yin20-1, duan20}, and the assumption of total ignorance found in \cite{xiao21}. To this end, this paper studies the following problem:

\begin{definition}[Adaptive Offline Policy Evaluation (AOPE)]
Adaptive Offline Policy Evaluation (AOPE) is Offline Policy Evaluation on $\script D = \{\tau_i \sim \mu^i\}_{i = 1}^n$, where $\mu^1, ..., \mu^n$ are chosen adaptively. That is, $\mu^i$ may depend on the trajectories $\tau_1, ... \tau_{i -1}$ (Figure \ref{Adaptive OPE versus OPE}).
\end{definition}

As opposed to vanilla OPE, the AOPE problem formulation allows for the data to have been collected according to a nearly arbitrary logging algorithm. When logging policies can be tuned according to previous trajectories, there is scope for starting from ``tabula rasa'', and iteratively refining the logging policy as we learn about the MDP. In other words, the logger can leverage online exploration techniques. Furthermore, by allowing arbitrary statistical dependence on previous trajectories, AOPE  addresses the key scenario of learning from intradependent, human-influenced datasets. 

The issue of defining an exploration assumption for an adaptive logger is an interesting one.  If $\mu^1, ... \mu^n$ were statically chosen, 
\begin{equation}\label{n-pol-dm}
    \bar d_m := \frac{1}{n}\min_{h, s, a}\sum_{i = 1}^nd^{\mu^i}_h(s, a) > 0
\end{equation} 
would be a good assumption. However, the quantity $\bar d_m$ as defined above is now a random variable.

We might assume that its \emph{expectation} is bounded below:
\begin{assumption}[Failed Exploration Assumption]\label{expected_exploration}
    For $\bar{d}_m > 0$, logging process $\script E$ satisfies a $\bar{d}_m$-\emph{expected}-exploration assumption if 
    \[\E[\frac{1}{n}\min_{h, s, a}\sum_{i = 1}^nd^{\mu^i}_h(s, a)] > \bar{d}_m\]
\end{assumption}
We will show by means of a lower bound (Theorem \ref{lower_bound}) that such an assumption does not allow us to derive high-probabulity bounds on estimation error.
We therefore find it most natural to levy our assumptions directly on the number of visitations to each $(h, s, a)$:

\begin{assumption}[Exploration Assumption]\label{assumption}
    For $\bar d_m > 0$, logging process $\script E$ satisfies a $(\bar d_m, \delta)$-exploration assumption if, with probability at least $1 - \delta$
    \[n_{h, s, a} > n\bar{d}_m\]
\end{assumption}

In the non-adaptive regime, $n_{h, s, a} \geq n\Bar{d}_m/2$ holds for Equation \ref{n-pol-dm}'s $\Bar{d}_m$ by a multiplicative Chernoff bound. This implies that our results hold for the single-logger setting as a special case, ensuring our work is a strict generalization of single-logger theory. Furthermore, this assumption is generally satisfied by reward-free exploration algorithms \citep{jin20rewardfree,qiao2022sample}.

Assumptions on the exploratory property of the logger will not always be necessary. Results that depend on Assumption \ref{assumption} will state it as a hypothesis.

We conclude this section by noting that, under Assumption \ref{assumption}, minimax bounds from the non-adaptive OPE setting can be easily recovered for AOPE in the following manner. For any dataset, $\script D$, let $N := \min_{h, s, a}n_{h, s, a}$ be the number of occurrences of the least-observed $(s_h, a_h)$ pair. If we consider a revised dataset, $\script D'$, that keeps only the first $N$ transitions out of $(s_h, a_h)$ for all $s_h, a_h$, we see that all transitions are now independent conditioned on $N$. Thus, the problem reduces to a generative-model type setting. In particular,  Theorem 3.7 in \cite{yin20-1} implies minimax-optimal offline learning for the adaptive case. However, we do not like throwing data away in this manner, and conjecture that it should not be necessary to do so. It would be preferable to obtain bounds that adapt to the quantities $\{n_{h, s, a}\}_{h, s, a}$ and the features of the MDP. To this end, we explore the extent to which \emph{instance-dependent} bounds on estimation error can be recovered in the adaptive setting.

\subsection{Our estimator}
We consider the TMIS (``Tabular Marginalized Importance Sampling'') estimator of $v^\pi$ studied in \citep{yin20-0}. This boils down to computing the value of a policy under the approximate MDP defined by $(\script S, \script A, \hat P, \hat r$, $\hat d_1$), with the estimators $\hat P$, $\hat r$ and $\hat d_1$ defined below.

That is, if $\script D = \{\tau_1, ..., \tau_n\}$, and $\tau_i = (s^{i}_1, a^i_1, r^i_1, ... s^i_H, a^i_H, r^i_H)$, we use plug-in estimates
\[\hat{P_h}(s' | s, a) = \frac{n_{h, s, a, s'}}{n_{h, s, a}} = \frac{1}{n_{h, s, a}}\sum_i\mathbf{1}_{\{s^i_h = s, a^i_h = a, s^i_{h + 1} = s'\}},\]\[
\hat r_h(s, a) = \frac{1}{n_{h, s, a}}\sum_{k = 1}^n r^k_h \mathbf{1}_{\{s^k_h = s, a^k_h = a\}},\]
subject to these quantities being well-defined ($n_{h, s, a} \neq 0$). If $n_{h, s, a} = 0$, we can define them to be $0$.

We also define $\hat{d}_1 := \hat d^\pi_1 := \frac{1}{n}\sum_{i = 1}^n e_{s^i_1}$ to be the plug-in estimate of $d_1$ computed from $\script D$ (where $e_j$ is the $j$th standard basis vector in $\R^S$).
 
We then let:
\[\hat{P}^\pi_h(s'|s) = \sum_a\pi_h(a | s)\hat{P_h}(s'|s, a),\]
\[\hat{r}^\pi_h(s) = \sum_a \pi_h(a | s)\hat{r}_h(s, a)\]
and iteratively compute $\hat d^\pi_h := \hat P^\pi_h \hat d^\pi_{h-1}$ for $h = 1, ... H$.

Finally, we form the estimate of value function as \[\hat{v}^\pi = \sum_{h = 1}^H \langle \hat d^\pi_h, \hat r^\pi_h \rangle.\]

\section{Theoretical Results}\label{theory_sec}

\subsection*{High-Probability Upper Bounds on Estimation Error}
We now turn our attention towards quantifying $\hat v^\pi$'s performance for AOPE. We first describe a high-probability, uniform error bound in terms of the number of visitations to each $(s_h, a_h)$ tuple. 

\begin{theorem}[High-probability uniform bound on estimation error in AOPE]\label{iduniformaope}
Suppose $\script D$ is a dataset conforming to AOPE, and $\hat v^\pi$ is formed using this dataset. Then, with probability at least $1 - \delta$,  the following holds for all deterministic policies $\pi$:
\[|\hat v^\pi - v^\pi| \leq \widetilde{O}\left(\sum_{h = 1}^H\sum_{s, a} H d^\pi_h(s, a)\sqrt{\frac{S}{n_{h, s, a}}}\right),\]
where $n_{h, s, a}$ is the number of occurrences of $(s_h, a_h)$ in $\script D$ and with the convention that $\frac{0}{0} = 0$.
\end{theorem}

This translates to the following worst-case bound, which underperforms the minimax-optimal bound (over deterministic policies) implied by \cite{yin20-1} by a factor of $\sqrt{H}$. Note that this result (and those that follow) use a $(\bar d_m, \delta/2)$-exploration assumption (Assumption \ref{assumption}).

\begin{corollary}[High-probability uniform bound on estimation error in AOPE] \label{uniformaope}Suppose that $\script D$, $\hat v^\pi$ are as in Theorem \ref{iduniformaope}. Suppose that, with probability $\geq 1 - \delta/2$, the logging process is such that $n_{h, s, a} \geq n\Bar{d}_m$ for all $h, s, a$. Then with probability $1 - \delta$, we have that
\[\sup_\pi|\hat v^\pi - v^\pi| \leq \widetilde{O}\left({H^2\sqrt{\frac{S}{n\bar d_m}}}\right).\]
\end{corollary}

We also give a high-probability, instance-dependent, \emph{pointwise} bound. In the pointwise case, we are able to shave off a $\sqrt{S}$ in the asymptotically dominant term. 

\begin{theorem}[Instance-dependent pointwise bound on estimation error in AOPE]\label{idpointwiseaope}
Fix a policy $\pi$, suppose $\script D$ is a dataset conforming to AOPE, and $\hat v^\pi$ is formed using this dataset. Assume that with probability $\geq 1 - \frac{\delta}{2}$, $n_{h, s, a} \geq n\bar d_m$ for all $h, s, a$, for some $\bar d_m > 0$. Then with probability at least $1-\delta$, we have:
\begin{align*}
&|\hat v^\pi - v^\pi| \leq  \widetilde{O}\left(\frac{H^3S}{n\bar d_m}\right)\\
&\quad +\widetilde{O}\left(\sum_{h = 1}^H \sum_{s, a} d^\pi_h(s, a)\sqrt{\frac{\Var_{s' \sim P_{h + 1}(\cdot | s, a)}[V^\pi_{h + 1}(s')]}{n_{h, s, a}}}\right).
\end{align*}
\end{theorem}

The above translates into the following worst-case bound, which is suboptimal by a factor of $\sqrt{H}$.

\begin{corollary}[Worst-case pointwise bound on estimation error in AOPE]\label{pointwiseaope}
Fix a policy $\pi$, suppose $\script D$ is a dataset conforming to AOPE, and $\hat v^\pi$ is formed using this dataset. Assume that with probability $\geq 1 - \frac{\delta}{2}$, $n_{h, s, a} \geq n\bar d_m$ for all $h, s, a$, for some $\bar d_m > 0$. Then with probability at least $1-\delta$, we have:
\[|\hat v^\pi - v^\pi| \leq \widetilde{O}\left({\sqrt{\frac{H^3}{n\bar d_m}}} + \frac{H^3S}{n\bar d_m}\right).\]
\end{corollary}

The form of Theorems \ref{iduniformaope} and \ref{idpointwiseaope} yield some key insights. For a given task, they explain which regions of the MDP are important by decomposing estimation error along $[H] \times \script S \times \script A$. For example, $\Var_{s' \sim P_{h + 1}(\cdot | s, a)}[V^\pi_{h + 1}(s')]$ is a measure of how ``pivotal'' a transition is -- that is, our uncertainty in the effect of the outcome of $(h, s, a)$ on our future reward. $d^\pi_h(s, a)$ is, of course, a measure of how likely the target policy is to visit $(h, s, a)$. Thus, these results tell us that it is important to design loggers that prioritize data-collection in regions of the MDP that are highly visited by $\pi$ or highly pivotal.

Though Corollary \ref{pointwiseaope} does not recover a minimax-optimal bound on the estimation error, $e^* := \widetilde{O}\left(\sqrt{\frac{H^2}{n\Bar{d}_m}}\right)$, Theorem \ref{idpointwiseaope} may be much tighter than $e^*$ for certain MDPs or certain policies. To take an extreme case, Theorem \ref{idpointwiseaope} shows that error decreases as $O(\frac{1}{n})$ when the MDP is deterministic.

\subsection*{Lower Bound for AOPE}
The following information-theoretic lower bound shows that for any learning algorithm, there is an adaptive logging process satisfying Assumption \ref{expected_exploration} such that error is constant with large probability.

\begin{theorem}\label{lower_bound}
    For any learning algorithm $L$ that takes as input a target policy and dataset, and outputs an estimate of the target policy's value, there exists a instance $I = (M, \pi, \script E)$ such that when $\script D \sim (M, \script E)$:
    \[\Pr[|L(\pi, \script D)] - v^\pi| > \frac{1}{2}] > \frac{1}{4}\]
    where $\script E$ is a logging process satisfying a $\bar d_m$-expected-exploration assumption, $M$ is an MDP, and $\pi$ is a policy.
\end{theorem}

This result tells us that assumptions on the expected exploration of an adaptive logger are not strong enough to rule out problematic pathologies. These pathologies inform the definition of Assumption \ref{assumption}, which is adopted in several results above.

\subsection{Proof sketches}

\subsection*{High-probability Results: \ref{iduniformaope}, \ref{uniformaope}, \ref{idpointwiseaope}, \ref{pointwiseaope}}
All of our results arise from applying concentration inequalities to functions of our estimates, $\hat{P}, \hat{r}, \hat{d}_0$. However, the first two quantities are formed using a mutually dependent dataset. The martingale structure of $|\hat{v}^\pi - v^\pi|$ used in \cite{yin20-1} is also lost in the adaptive setting, so there is no straightforward way to apply concentration. However, the tape model tells us that for fixed $h, s, a$ and $n_{h, s, a}$, the error admits an expression amenable to concentration. Therefore, by using a covering argument over $\{n_{h, s, a}\}$, we are able to obtain our bounds. Appendix \ref{tape} covers these details more carefully.

The proof of Theorem \ref{iduniformaope} follows by a simulation lemma-type expansion of the error, which leads to a dominant term of the form $\sum_h \E_{s_h, a_h \sim \pi, \script M}[(\hat P_{h + 1}(\cdot | s_h, a_h) - P_{h + 1}(\cdot | s_h, a_h))^T \hat V^\pi_{h + 1}]$, and smaller terms governed by $\hat r$ and $\hat d_1$. As mentioned above, we must cover all possible number of occurrences of each $(s_h, a_h)$ across trajectories while applying concentration, leading to the $HSAn$ term inside the logarithm.  The full proof is deferred to Appendix \ref{uniform aope proof}.

Inspired by \cite{azar17}, Theorem \ref{idpointwiseaope} is proved by applying concentration inequalities (with the same covering trick as Theorem \ref{iduniformaope}) to $(\hat P_{h +1} - P_{h + 1})V^\pi_{h + 1}$ and $(\hat P_{h +1} - P_{h + 1})(\hat V^\pi_{h + 1} - V^\pi_{h + 1})$ separately, instead of $(\hat P_{h +1} - P_{h + 1})\hat V^\pi_{h + 1}$. In order to treat the dominant term, we use Bernstein's inequality. The residual term scales with $\frac{1}{n} \ll \frac{1}{\sqrt{n}}$, which allows us to use cruder bounds when treating it. To recover the worst-case bound in the corollary, we analyze the variance term with the canonical equality $\sum_h E_\pi[\Var_{s' \sim P_h(\cdot | s, a)}[V^\pi_h(s')]] \leq \Var_\pi[\sum_h r_h] \leq H^2$. The full proof is presented in Appendix \ref{pointwise aope proof}.

\subsection{Lower Bound: Theorem \ref{lower_bound}}
The lower bound is obtained by a packing argument over tree-shaped MDPs with an adaptive logger that does good exploration with probability $\frac{1}{2}$ and bad exploration with probability $\frac{1}{2}$. The full proof can be found in Appendix \ref{lower_appendix}.
\section{Empirical Results}\label{simulations}

\subsection{Experimental Motivation and Design}
Our theoretical results certify that the TMIS estimator achieves low error even with adaptively logged data. However, they leave open some important questions regarding the behavior of TMIS estimation under adaptive data:
\begin{enumerate}
    \item In multi-arm bandit literature, it has been established \citep{Shin19Bias} that optimistic exploration causes negative bias in sample means. Does this behavior emerge in RL as well? 
    \item Our results hold for arbitrary adaptivity, which may be adversarially chosen. But is it possible that some forms of adaptivity are beneficial?  \footnote{Note that a negative answer here does not contradict our discourse on the importance of adaptivity in Offline RL, which is that it allows us to choose logging policies more wisely. From a practical point of view, we are making an ``unfair'' comparison between a realistic dataset, which is the result of earnest exploration of the MDP, and an unrealistic dataset, which we provide with the resultant suite of good exploration policies at the outset.}
\end{enumerate}
In this section, we probe these questions empirically. On the way, we validate our theoretical  results. As our adaptive logger, we use UCB-VI. We gather data using UCB-VI, and then roll out an independent ``Shadow'' dataset using the same policies.  For details on the data-collection method, see section \ref{simulations appendix} of the Appendix. As an optimistic algorithm, UCB-VI is well-suited to testing Question 1. Furthermore, as UCB-VI steers the data-collection procedure towards high-value states, it is conceivable that our estimator will benefit from the adaptivity for optimal $\pi$. UCB-VI can ``react'' to unwanted outcomes in trajectory $\tau_i$, where the Shadow dataset cannot.

\subsection{Results}\label{emp results}

\begin{figure}[ht]
\vskip 0.2in
\begin{center}
\centerline{\includegraphics[width=\columnwidth]{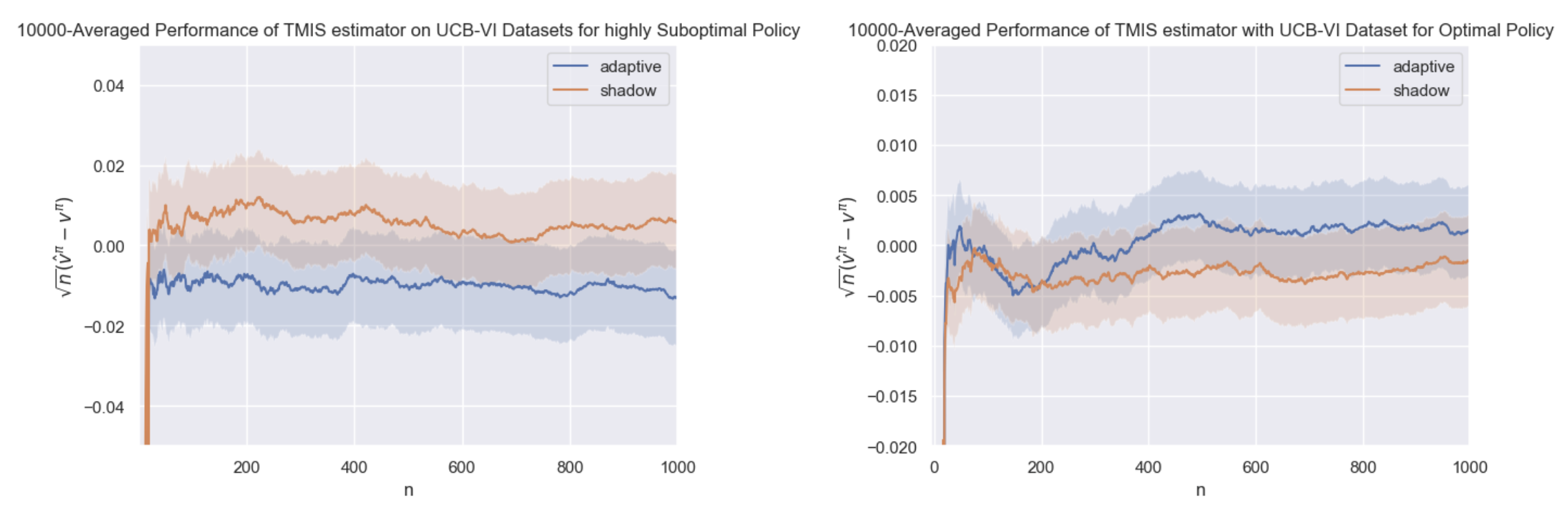}}
\caption{For different $\pi$, the blue curves show the average value of $\sqrt{n}\times(\hat{v}^\pi - v^\pi)$, where $\hat{v}^\pi$ is computed on the first $n \leq N$ trajectories of each of the $10,000$ adaptive datasets. The orange curves are computed in the same way, except using the $10,000$ shadow datasets. On the lefthand side, $\pi$ is very suboptimal. On the righthand side, $\pi$ is optimal. Confidence intervals are 95\% Gaussian.}
\label{plots}
\end{center}
\vskip -0.2in
\end{figure}

We first consider a highly sub-optimal target policy. The lefthand side of Figure \ref{plots} shows two curves. Each curve plots the $\sqrt{n}$-scaled estimation error $\sqrt{n}(\hat{v}^\pi - v^\pi)$ against $n$, averaged over $10,000$ runs of the data-collection process ($10,000$ runs of UCB-VI and each run's corresponding shadow dataset). For each $n$ and for each curve, this average is computed with respect to the first $n \leq N$ trajectories in each dataset. Theorem \ref{uniformaope} tells us that these quantities will live in a band around zero, but does not give us information on their sign. 

Plotting the $95\%$ confidence interval around each curve, there appears to be a distinction between the signed behaviors of the estimator for adaptive data vs non-adaptive data, though both confidence intervals cover $0$. This suggests that there are measurable differences in the signed behavior of $\hat{v}^\pi$ when adaptive data is used, even if the logging policies are the same!

The righthand side of Figure \ref{plots} plots the same quantities, but considers a high-value target policy instead of a low-value target policy. The curves have switched positions in this plot, but our confidence intervals suggest that this distinction may not be meaningful.

On the whole, it seems that UCB-VI leads to negative bias in our estimates (especially for suboptimal policies), but limitations on the computational resources available to us constrain us to showing weak evidence of this conjecture. We also note that the magnitude of the error is indistinguishable for adaptive and non-adaptive data, giving us hope that tighter instance-dependent AOPE bounds can be obtained. We also note that these simulations act as a good sanity check of our result, by showing that $\sqrt{n}|\hat{v}^\pi - v^\pi|$ does not explode.

In summary, based on these simulations, we posit that AOPE is not inherently more difficult than OPE with non-adaptive data. We find weak evidence of an affirmative answer to the first experimental question that we posed, and no evidence of an affirmative answer to the second.

\section{Conclusion and Future Work}
By divorcing itself from i.i.d. assumptions, offline RL theory can tie its fortunes to those of the broader RL theory community. Provable means of exploring MDPs \citep{azar17, jin20rewardfree} underscore the benefits of understanding how to learn from adaptive datasets.

In this paper, we have shown instance-dependent bounds on the estimation error of a very natural estimator under adaptively collected data. These bounds reveal the individual contribution of each $(h, s, a)$ to the error, based on fundamental properties of the MDP and target policy. Independently to this, we have noticed that, under reasonable exploration assumptions, minimax-rates from the non-adaptive setting can be directly ported into the adaptive setting. The disconnect between the worst-case bounds implied by our instance-dependent results and the minimax-optimal bound is intriguing. It arises from the existence of a martingale structure across  timesteps in non-adaptive data \citep{yin20-1}; this structure disappears in the adaptive regime. Bridging this gap is an interesting avenue for future work, as it would fully allay lingering doubts about the efficiency of adaptive data. Even so, the Martingale argument yields a rather less pithy expression than Theorem \ref{idpointwiseaope}, and so we are not overly bothered by the discrepancy. 

Though, at first glance, the form of our worst-case bounds on the sample complexity look like a looser version of those obtained in the iid case, we emphasize that in such results, the $d_m$ term is doing implausibly heavy lifting. \cite{xiao21} demonstrates that the optimistic assumption (made by many authors) that $d_m \approx \frac{1}{SA}$ essentially presupposes a huge amount of prior knowledge on the MDP. Realistically, one is hard-pressed to do non-adaptive logging without $d_m \ll \frac{1}{SA}$. On the other hand, in our work, $\bar d_m$  represents the realized quality of an adaptive logger’s exploration, which should be good for e.g. reward-free logging. The upshot is that in the setting where we do not have pre-existing knowledge on the MDP, sample-complexity results for adaptive logging are better. Thus, restricting our attention to practically plausible loggers, our results promise better sample complexity. Orthogonal to such concerns, in virtually any real-world dataset, adaptivity will be present. Thus, our bounds are important even in the pathological cases where they underperform our comparators.

We also began the process of exploring the effects of various types of adaptivity on our estimates. We find that, even controlling for identical logging policies, there are differences in the behavior of the TMIS estimator for OPE and AOPE. Systematically understanding how datasets generated by different exploration algorithms interact with existing OPE estimators is yet more fertile ground for research.

 We have confined our attention to the microcosmic space of tabular MDPs. Though generalizing our results to wider classes of MDPs is important future work, the expressions derived here can and should inform data-collection in much more complicated environments. In theoretically clarifying the extent to which OPE methods carry over to the AOPE setting, we hope we have moved towards bridging the large gap that exists between theory and practice in offline RL. 
\nocite{*}

\bibliography{citations.bib}

\begin{thebibliography}{33}
\providecommand{\natexlab}[1]{#1}
\providecommand{\url}[1]{\texttt{#1}}
\expandafter\ifx\csname urlstyle\endcsname\relax
  \providecommand{\doi}[1]{doi: #1}\else
  \providecommand{\doi}{doi: \begingroup \urlstyle{rm}\Url}\fi

\bibitem[Azar et~al.(2017)Azar, Osband, and Munos]{azar17}
Azar, M.~G., Osband, I., and Munos, R.
\newblock Minimax regret bounds for reinforcement learning.
\newblock In \emph{International Conference on Machine Learning}, pp.\
  263--272. PMLR, 2017.

\bibitem[Cai et~al.(2017)Cai, Ren, Zhang, Malialis, Wang, Yu, and
  Guo]{cai2017real}
Cai, H., Ren, K., Zhang, W., Malialis, K., Wang, J., Yu, Y., and Guo, D.
\newblock Real-time bidding by reinforcement learning in display advertising.
\newblock In \emph{Proceedings of the Tenth ACM International Conference on Web
  Search and Data Mining}, pp.\  661--670, 2017.

\bibitem[Dann et~al.(2017)Dann, Lattimore, and Brunskill]{dann2017unifying}
Dann, C., Lattimore, T., and Brunskill, E.
\newblock Unifying pac and regret: Uniform pac bounds for episodic
  reinforcement learning.
\newblock \emph{Advances in Neural Information Processing Systems}, 30, 2017.

\bibitem[Duan et~al.(2020)Duan, Jia, and Wang]{duan20}
Duan, Y., Jia, Z., and Wang, M.
\newblock Minimax-optimal off-policy evaluation with linear function
  approximation.
\newblock In \emph{International Conference on Machine Learning}, pp.\
  2701--2709. PMLR, 2020.

\bibitem[Dudik et~al.(2011)Dudik, Langford, and Li]{dudik2011doubly}
Dudik, M., Langford, J., and Li, L.
\newblock Doubly robust policy evaluation and learning, 2011.

\bibitem[Jiang \& Li(2016)Jiang and Li]{jiang2016doubly}
Jiang, N. and Li, L.
\newblock Doubly robust off-policy value evaluation for reinforcement learning,
  2016.

\bibitem[Jin et~al.(2020)Jin, Krishnamurthy, Simchowitz, and
  Yu]{jin20rewardfree}
Jin, C., Krishnamurthy, A., Simchowitz, M., and Yu, T.
\newblock Reward-free exploration for reinforcement learning.
\newblock In \emph{International Conference on Machine Learning, {ICML} 2020},
  volume 119, pp.\  4870--4879. {PMLR}, 2020.

\bibitem[Jin et~al.(2021)Jin, Yang, and Wang]{jin2021pessimism}
Jin, Y., Yang, Z., and Wang, Z.
\newblock Is pessimism provably efficient for offline rl?
\newblock In \emph{International Conference on Machine Learning}, pp.\
  5084--5096. PMLR, 2021.

\bibitem[Jin et~al.(2022)Jin, Ren, Yang, and Wang]{Jin22Overlap}
Jin, Y., Ren, Z., Yang, Z., and Wang, Z.
\newblock Policy learning "without" overlap: Pessimism and generalized
  empirical bernstein's inequality.
\newblock \emph{CoRR}, abs/2212.09900, 2022.
\newblock \doi{10.48550/arXiv.2212.09900}.
\newblock URL \url{https://doi.org/10.48550/arXiv.2212.09900}.

\bibitem[Kallus et~al.(2020)Kallus, Saito, and Uehara]{kallus}
Kallus, N., Saito, Y., and Uehara, M.
\newblock Optimal off-policy evaluation from multiple logging policies, 2020.
\newblock URL \url{https://arxiv.org/abs/2010.11002}.

\bibitem[Kiran et~al.(2020)Kiran, Sobh, Talpaert, Mannion, Sallab, Yogamani,
  and Pérez]{kivan_drive}
Kiran, B.~R., Sobh, I., Talpaert, V., Mannion, P., Sallab, A. A.~A., Yogamani,
  S., and Pérez, P.
\newblock Deep reinforcement learning for autonomous driving: A survey, 2020.
\newblock URL \url{https://arxiv.org/abs/2002.00444}.

\bibitem[Lambert et~al.(2022)Lambert, Wulfmeier, Whitney, Byravan, Bloesch,
  Dasagi, Hertweck, and Riedmiller]{Lambert22Challenges}
Lambert, N., Wulfmeier, M., Whitney, W.~F., Byravan, A., Bloesch, M., Dasagi,
  V., Hertweck, T., and Riedmiller, M.~A.
\newblock The challenges of exploration for offline reinforcement learning.
\newblock \emph{CoRR}, abs/2201.11861, 2022.

\bibitem[Lange et~al.(2012)Lange, Gabel, and Riedmiller]{lange2012batch}
Lange, S., Gabel, T., and Riedmiller, M.
\newblock Batch reinforcement learning.
\newblock In \emph{Reinforcement learning}, pp.\  45--73. Springer, 2012.

\bibitem[Le et~al.(2019)Le, Voloshin, and Yue]{Le19_constraints}
Le, H.~M., Voloshin, C., and Yue, Y.
\newblock Batch policy learning under constraints.
\newblock In Chaudhuri, K. and Salakhutdinov, R. (eds.), \emph{International
  Conference on Machine Learning, {ICML} 2019}, volume~97, pp.\  3703--3712.
  {PMLR}, 2019.

\bibitem[Mou et~al.(2022)Mou, Wainwright, and Bartlett]{relatedwork}
Mou, W., Wainwright, M.~J., and Bartlett, P.~L.
\newblock Off-policy estimation of linear functionals: Non-asymptotic theory
  for semi-parametric efficiency, 2022.
\newblock URL \url{https://arxiv.org/abs/2209.13075}.

\bibitem[Qiao \& Wang(2022{\natexlab{a}})Qiao and Wang]{qiao2022near}
Qiao, D. and Wang, Y.-X.
\newblock Near-optimal deployment efficiency in reward-free reinforcement
  learning with linear function approximation.
\newblock \emph{arXiv preprint arXiv:2210.00701}, 2022{\natexlab{a}}.

\bibitem[Qiao \& Wang(2022{\natexlab{b}})Qiao and Wang]{qiao2022offline}
Qiao, D. and Wang, Y.-X.
\newblock Offline reinforcement learning with differential privacy.
\newblock \emph{arXiv preprint arXiv:2206.00810}, 2022{\natexlab{b}}.

\bibitem[Qiao et~al.(2022)Qiao, Yin, Min, and Wang]{qiao2022sample}
Qiao, D., Yin, M., Min, M., and Wang, Y.-X.
\newblock Sample-efficient reinforcement learning with loglog({T}) switching
  cost.
\newblock In \emph{International Conference on Machine Learning}, pp.\
  18031--18061. PMLR, 2022.

\bibitem[Raghu et~al.(2017)Raghu, Komorowski, Celi, Szolovits, and
  Ghassemi]{raghu2017continuous}
Raghu, A., Komorowski, M., Celi, L.~A., Szolovits, P., and Ghassemi, M.
\newblock Continuous state-space models for optimal sepsis treatment: a deep
  reinforcement learning approach.
\newblock In \emph{Machine Learning for Healthcare Conference}, pp.\  147--163,
  2017.

\bibitem[Raghu et~al.(2018)Raghu, Gottesman, Liu, Komorowski, Faisal,
  Doshi{-}Velez, and Brunskill]{Raghu19matters}
Raghu, A., Gottesman, O., Liu, Y., Komorowski, M., Faisal, A., Doshi{-}Velez,
  F., and Brunskill, E.
\newblock Behaviour policy estimation in off-policy policy evaluation:
  Calibration matters.
\newblock \emph{CoRR}, abs/1807.01066, 2018.
\newblock URL \url{http://arxiv.org/abs/1807.01066}.

\bibitem[Schmucker et~al.(2021)Schmucker, Wang, Hu, and Mitchell]{shmucker_ed}
Schmucker, R., Wang, J., Hu, S., and Mitchell, T.~M.
\newblock Assessing the performance of online students -- new data, new
  approaches, improved accuracy, 2021.
\newblock URL \url{https://arxiv.org/abs/2109.01753}.

\bibitem[Shin et~al.(2019)Shin, Ramdas, and Rinaldo]{Shin19Bias}
Shin, J., Ramdas, A., and Rinaldo, A.
\newblock Are sample means in multi-armed bandits positively or negatively
  biased?
\newblock In Wallach, H.~M., Larochelle, H., Beygelzimer, A.,
  d'Alch{\'{e}}{-}Buc, F., Fox, E.~B., and Garnett, R. (eds.), \emph{Advances
  in Neural Information Processing Systems 32: Annual Conference on Neural
  Information Processing Systems 2019, NeurIPS 2019, December 8-14, 2019,
  Vancouver, BC, Canada}, pp.\  7100--7109, 2019.

\bibitem[Thomas \& Brunskill(2016)Thomas and
  Brunskill]{thomas2016dataefficient}
Thomas, P.~S. and Brunskill, E.
\newblock Data-efficient off-policy policy evaluation for reinforcement
  learning, 2016.

\bibitem[Wang et~al.(2020)Wang, Du, Yang, and Salakhutdinov]{wang2020reward}
Wang, R., Du, S.~S., Yang, L., and Salakhutdinov, R.~R.
\newblock On reward-free reinforcement learning with linear function
  approximation.
\newblock \emph{Advances in neural information processing systems},
  33:\penalty0 17816--17826, 2020.

\bibitem[Wang et~al.(2017)Wang, Agarwal, and Dudik]{wang2017optimal}
Wang, Y.-X., Agarwal, A., and Dudik, M.
\newblock Optimal and adaptive off-policy evaluation in contextual bandits,
  2017.

\bibitem[Xiao et~al.(2022)Xiao, Lee, Dai, Schuurmans, and Szepesvari]{xiao21}
Xiao, C., Lee, I., Dai, B., Schuurmans, D., and Szepesvari, C.
\newblock The curse of passive data collection in batch reinforcement learning.
\newblock In \emph{International Conference on Artificial Intelligence and
  Statistics}, pp.\  8413--8438. PMLR, 2022.

\bibitem[Xie \& Jiang(2021)Xie and Jiang]{Xie20batch}
Xie, T. and Jiang, N.
\newblock Batch value-function approximation with only realizability.
\newblock In Meila, M. and Zhang, T. (eds.), \emph{International Conference on
  Machine Learning, {ICML}}, volume 139 of \emph{Proceedings of Machine
  Learning Research}, pp.\  11404--11413. {PMLR}, 2021.

\bibitem[Xie et~al.(2019)Xie, Ma, and Wang]{xie2019towards}
Xie, T., Ma, Y., and Wang, Y.-X.
\newblock Towards optimal off-policy evaluation for reinforcement learning with
  marginalized importance sampling.
\newblock \emph{Advances in Neural Information Processing Systems}, 32, 2019.

\bibitem[Yin \& Wang(2020)Yin and Wang]{yin20-0}
Yin, M. and Wang, Y.-X.
\newblock Asymptotically efficient off-policy evaluation for tabular
  reinforcement learning.
\newblock In \emph{International Conference on Artificial Intelligence and
  Statistics}, pp.\  3948--3958. PMLR, 2020.

\bibitem[Yin et~al.(2021)Yin, Bai, and Wang]{yin20-1}
Yin, M., Bai, Y., and Wang, Y.-X.
\newblock Near-optimal provable uniform convergence in offline policy
  evaluation for reinforcement learning.
\newblock In \emph{International Conference on Artificial Intelligence and
  Statistics}, pp.\  1567--1575. PMLR, 2021.

\bibitem[Yin et~al.(2022)Yin, Duan, Wang, and Wang]{yin2022near}
Yin, M., Duan, Y., Wang, M., and Wang, Y.-X.
\newblock Near-optimal offline reinforcement learning with linear
  representation: Leveraging variance information with pessimism.
\newblock \emph{arXiv preprint arXiv:2203.05804}, 2022.

\bibitem[Yu et~al.(2019)Yu, Liu, and Nemati]{Yu_med}
Yu, C., Liu, J., and Nemati, S.
\newblock Reinforcement learning in healthcare: A survey, 2019.
\newblock URL \url{https://arxiv.org/abs/1908.08796}.

\bibitem[Zhang et~al.(2022)Zhang, Zhang, Ni, and Wang]{zhang2022off}
Zhang, R., Zhang, X., Ni, C., and Wang, M.
\newblock Off-policy fitted q-evaluation with differentiable function
  approximators: Z-estimation and inference theory.
\newblock \emph{arXiv preprint arXiv:2202.04970}, 2022.

\end{thebibliography}
\bibliographystyle{icml2023}
\newpage
\section*{Appendix}
\section{Concentration inequalities}

\begin{lemma}[Hoeffding's Inequality] Let $x_1, ..., x_n$ be bounded random variables such that $E[x_i] = 0$ and $|x_i| \leq \xi_i$ with probability 1. Then for any
$\epsilon>0$ we have:
\[\Pr[|\frac{1}{n}\sum_{i = 1}^nx_i| > \epsilon] \leq \exp\{-2\epsilon^2n^2/\sum_{i=1}^n \xi_i^2\}.\]

\end{lemma}
\begin{lemma}[Bernstein's Inequality]
Let $x_1, ..., x_n$ be independent bounded random variables such that $E[x_i] = 0$ and $|x_i| \leq \xi$ with probability 1. Let $\sigma^2 = \frac{1}{n}\sum_{i = 1}^n \Var[x_i]$, then with probability $1 - \delta$ we have:
\[\frac{1}{n}\sum_{i = 1}^nx_i \leq \sqrt{\frac{\sigma^2\log\frac{1}{\delta}}{n}} + \frac{2\xi}{3n}\log\frac{1}{\delta}.\]
\end{lemma}

\begin{lemma}[d-dimensional Concentration]\label{d-dim conc}%
Let $z$ be a discrete random variable taking values in $\{1, ... d\}$. Let $q$ be the associated pmf. Assume we have $n$ i.i.d. samples $z_1, ... z_n$, and define $\hat{q}$ by $\hat{q}_j = \frac{1}{n}\sum_{i = 1}^n \mathbf{1}_{\{z_i = j\}}$. Then for any $\epsilon > 0$,
\[\Pr[\|\hat{q} - q\|_1 \geq \sqrt{d}(\frac{1}{\sqrt{n}} + \epsilon)] \leq \exp\{-N\epsilon^2\}.\]
\end{lemma}

\section{The ``Tape'' View of Adaptivity}\label{tape}
The simulation lemma allows us to decompose the error $\hat{v}^\pi - v^\pi$ into error terms corresponding to each possible $(h, s, a)$-tuple, $E_{h, s, a}$. $E_{h, s, a}$ will be instantiated differently across this paper, but it will always be a (data-independent) function, $f$, of the observed transitions out of $(h, s, a)$, $\{s^{i}_{h + 1}\}_{i: s^i_h = s, a^i_h = a}$. We will therefore be interested in the following type of expression:
\[E_{h, s, a} = \frac{1}{n_{h, s, a}}\sum_{i : s^i_h = s, a^i_h = a} f(s^i_{h + 1}) - \E[f(s') | s' \sim P_{h + 1}(\cdot | s, a)].\]
We want to apply concentration inequalities to each $E_{h, s, a}$. We need to be careful though, because $\{s^i_{h + 1}\}_{i = 1}^n$ are mutually dependent. 
The crucial point in bounding $E_{h, s, a}$ is to observe that for fixed $n_{h, s, a}$, $E_{h, s, a}$ is amenable to concentration. More formally, we write:
\[E_{h, s, a} = \sum_{j = 1}^n \mathbf{1}_{\{n_{h, s, a} = j\}} \frac{1}{n_{h, s, a}}\sum_{i : s^i_h = s, a^i_h = a} f(s^i_{h + 1}) - \E[f(s') | s' \sim P_{h + 1}(\cdot | s, a)]\]
\[=\sum_{j = 1}^n \mathbf{1}_{\{n_{h, s, a} = j\}} \underbrace{\frac{1}{j}\sum_{i = 1}^j f(s^{(i)}_{h + 1}) - \E[f(s') | s' \sim P_{h + 1}(\cdot | s, a)]}_{x_j}.\]
where $(i)$ re-indexes into trajectories that visit $(s_h, a_h)$. Now each $f(s^{(i)}_{h + 1})$ is independently distributed according to $f(s')$ with $s' \sim P_h(\cdot | s, a)$. 

In order to control $E_{h, s, a}$ with probabiliy $\delta$, then, it suffices to control each $x_j$ with probability $\delta/n$. In this way, covering $n_{h, s, a}$ for all $h, s, a$ yields bounds for the adaptive case. Note that in this paper the codomain of $f$ will be either $\R$ or $\R^S$.

One way to visualize the dependence structure in $\script D$ that we have leveraged above is to imagine a machine with a tape for each $(h, s, a)$. The tape $(h, s, a)$ contains an infinite string of i.i.d draws of $s' \sim P_h(\cdot | s, a)$, sampled before the logging begins. Every time $(h, s, a)$ is visited by the logging algorithm, the $s'$  read off of the corresponding tape, and the tape is advanced. Thus, a ``frontier'' is defined by $\{n_{h, s, a}\}_{h, s, a \in H \times \mathcal{S} \times \mathcal{A}}$ according to the logger's behavior. But, crucially, for any $n_{h, s, a}$, the transitions sampled off of the tape are i.i.d. This is the observation that justifies the application of concentration inequalities on $x_j$.

Whenever concentration is applied in this paper, the above covering argument is needed. Henceforth, every concentration argument will implicitly invoke the above covering trick, thus incurring a factor of $n$ within the log term.

\begin{figure}
\centering
  \includegraphics[width=\linewidth]{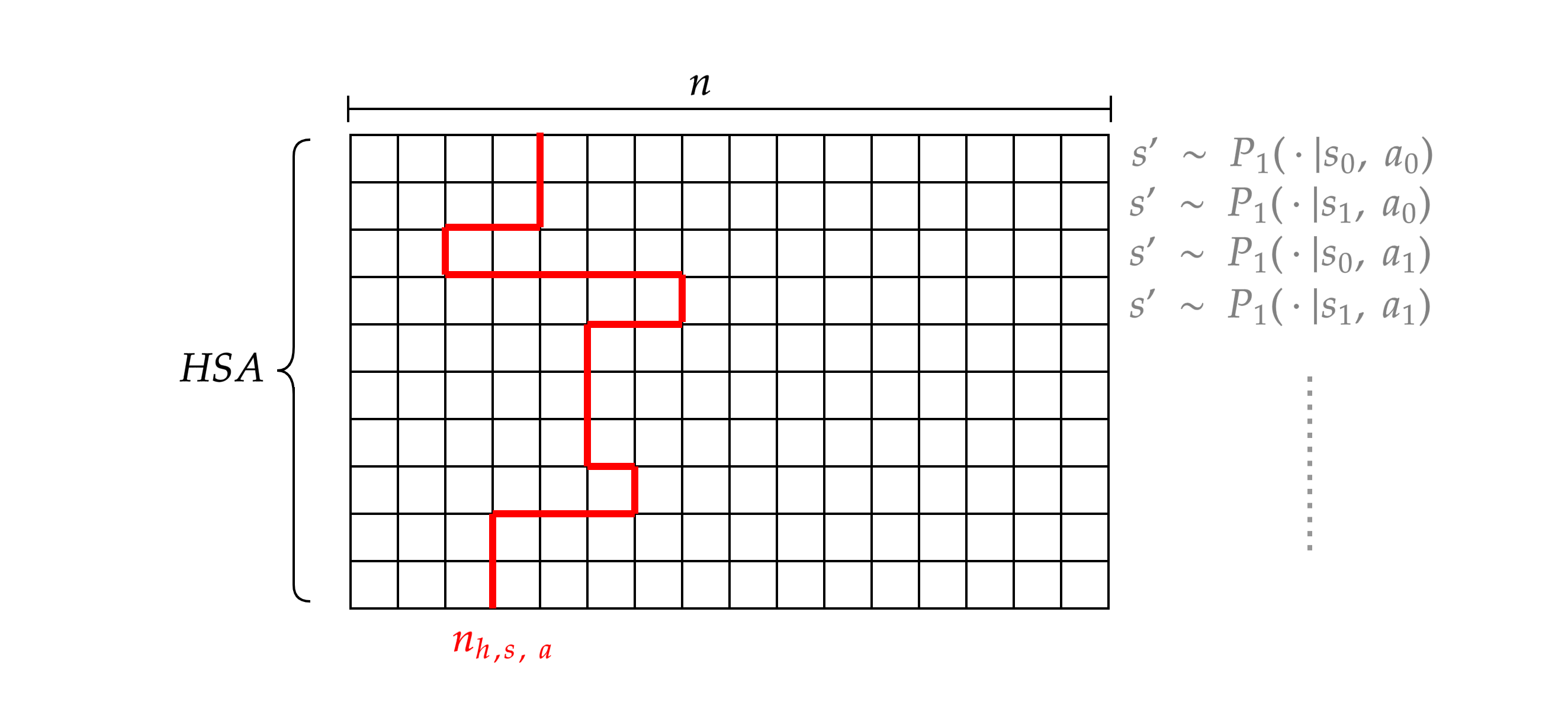}
\caption{A reproduction of Figure \ref{tape pic}. Each row $(h, s, a)$ should be thought to contain $n$ i.i.d. samples from $P_{h + 1}(\cdot | s, a)$. The red ``frontier'' tracks how many samples have, in fact, been used by the logger (this quantity is always bounded by $n$). }
\end{figure}

\section{Preliminaries and Proof of Uniform Bounds on $|\hat{v}^\pi - v^\pi|$}\label{uniform aope proof}
We assume for convenience that $r$ and $d_0$ (the reward function and the initial distribution) are \emph{known} in Appendices \ref{uniform aope proof} and \ref{pointwise aope proof}. Because the error of $\hat{v}^\pi$ is dominated by error due to $\hat{P}$, this assumption changes only constants. For further discussion of this point see Appendix \ref{wlog_sec}.

By the Simulation Lemma \citep{dann2017unifying}, we have the following representation of the error, $|\hat{v}^\pi - v^\pi|$:
\begin{equation}\label{sim_lemma}
    |\hat{v}^\pi - v^\pi| \leq \sum_{h = 1}^{H-1}|\langle d^\pi_h, (\hat{P}_{h + 1} - P_{h +1})\hat{V}^\pi_{h + 1}\rangle| \leq \sum_{h = 1}^{H-1}\sum_{s, a}d^\pi_h(s, a)|(\hat{P}_{h + 1}(\cdot | s, a) - P_{h + 1}(\cdot | s, a))^T\hat{V}^\pi_{h + 1}(\cdot)|,
\end{equation}
where we remind the reader that $\hat P_t$ is the estimated transition kernel moving into the $t$-th timestep, and $\hat V^\pi_t$ is the time-$t$ value function for $\pi$ under the approximate MDP defined by $\{\hat P_t\}_t$. Finally $d^\pi_t \in \R^{S\times A}$ is the marginal distribution of $(s_t, a_t)$ under the true MDP and $\pi$. Theorem \ref{iduniformaope} follows quite directly from the above expansion, but we need to cover $n_{h, s, a}$ to take care of the dependence induced by the adaptivity. 

In contrast to \citep{yin20-0, yin20-1}, our exploration assumption is levied on $\{n_{h, s, a}\}$ directly, rather than on the exploration parameters of the logging policy/ies (which would not sufficiently characterize the exploratory properties of an adaptive dataset). Thus, for each proof, we take care to budget for $\delta/2$ total failure probability across our concentration arguments -- the remaining $\delta/2$ accounts for the failure case where $n_{h, s, a} = 0$ for some $h, s, a$. 

\subsection{Proof of Theorem \ref{iduniformaope}  and Corollary \ref{uniformaope}}\label{uniform aope proof}

Using the third expression in equation \ref{sim_lemma}, Holder's inequality, and concentration in 1-norm (Lemma \ref{d-dim conc}) we have (with probability at least $1 - \delta/2$):
\[|\hat{v}^\pi - v^\pi| \leq \sum_{h = 1}^{H-1}\sum_{s, a}d^\pi_h(s, a) \|\hat{P}_{h + 1}(\cdot | s, a) - P_{h + 1}(\cdot | s, a)\|_1 \|\hat{V}^\pi_{h + 1}(\cdot)\|_\infty \]\[\leq\sum_{h = 1}^{H-1}\sum_{s, a} d^\pi_h(s, a)H(\sqrt{\frac{S\log{\frac{HSAn}{\delta}}}{n_{h, s, a}}} + \sqrt{\frac{S}{n_{h, s, a}}}) \leq 2\sum_{h = 1}^{H-1}\sum_{s, a}d^\pi_h(s, a)H\sqrt{\frac{S\log{\frac{HSAn}{\delta}}}{n_{h, s, a}}},\]
where it must be noted that we have made the near-vacuous assumption that $n \geq e\delta \geq e\delta/HSA$ for which it suffices for $n$ to be at least 3.

Note that all bounding was done independently of the policy $\pi$. Therefore this result holds uniformly across all policies. 

Corollary \ref{uniformaope} follows immediately.

\section{Proof of Pointwise Bounds on $|\hat{v}^\pi - v^\pi|$}\label{pointwise aope proof}
We now derive a slightly more sophisticated pointwise bound on the estimation error. We achieve this by replacing the estimated value function with the true value function in the simulation lemma, and pushing the residual into a lower-order term. 

Again, we budget for $\delta/2$ failure probability, which provides for $n_{h, s, a} > 0$ for all $h, s, a$ in our proofs.

\subsection{Proof of Theorem \ref{idpointwiseaope}}
Again, we start with Equation (\ref{sim_lemma}), which we re-express as
\[|\hat{v}^\pi\ - v^\pi| \leq \sum_{h = 1}^{H-1}\sum_{s, a}|d^\pi_h(s, a)[\underbrace{(\hat{P}_{h + 1}(\cdot | s, a) - P_{h + 1}(\cdot | s, a))^TV^\pi_{h + 1}(\cdot)}_{(*)} + \underbrace{(\hat{P}_{h + 1}(\cdot | s, a) - P_h(\cdot | s, a))^T(\hat{V}^\pi_{h + 1} - V^\pi_{h + 1})(\cdot)}_{(**)}]|.\]

\subsection{Bounding (*)}
For arbitrary $h, s, a$, notice that 
\[|(\hat{P}_{h + 1}(\cdot | s, a) - P_{h + 1}(\cdot | s, a))^TV^\pi_{h + 1}(\cdot)| = |\sum_{s'}(\hat{P}_{h + 1}(s' | s, a) - P_{h + 1}(s' | s, a))^TV^\pi_{h + 1}(s')| \]
\[
=|\sum_{s'}(\frac{1}{n_{h, s, a}}\sum_{i: s^i_h = s, a^i_h = a}\mathbf{1}_{\{s^i_{h + 1} = s'\}} - P_{h+1}(s' | s, a))V^\pi_{h + 1}(s')|\]\[ \leq \frac{1}{n_{h, s, a}}|\sum_{i : s^i_h = s, a^i_h = a} V^\pi_{h + 1}(s^i_{h + 1}) - \E_{s^\prime \sim P_{h + 1}(\cdot | s, a)}[V^\pi_{h + 1}(s^\prime)]|.\]

Using the tape-argument from Appendix \ref{tape}, we apply Bernstein's inequality while covering $n_{h, s, a}$ to get that with probability at least $1 - \delta/4HSAn$:
\begin{equation}|(\hat{P}_{h + 1}(\cdot | s, a) - P_{h + 1}(\cdot | s, a))^TV^\pi_{h + 1}(\cdot)| \leq \sqrt{\frac{2\Var[V^\pi_{h + 1}(s') | s' \sim P_{h + 1}(\cdot | s, a)]\log{\frac{4HSAn}{\delta}}}{n_{h, s, a}}} + \frac{4H}{3n_{h, s, a}}\log{\frac{4HSAn}{\delta}}.\end{equation}

\subsection{Bounding (**)}
By Holder's inequality:
\[(**) \leq \|\hat{P}_{h + 1}(\cdot | s, a) - P_{h + 1}(\cdot | s, a)\|_1\|(\hat{V}^\pi_{h + 1} - V^\pi_{h + 1})(\cdot)\|_\infty\]

The first term above is bounded with the same $S$-dimensional concentration inequality that we used in \ref{uniform aope proof}, Lemma \ref{d-dim conc}. With probability $1 - \delta/4HSA$, the following holds.
 
\begin{equation}\label{1-norm control} \|\hat{P}_{h + 1}(\cdot | s, a) - P_{h + 1}(\cdot | s, a)\|_1 \leq \sqrt{\frac{S\log{\frac{4HSAn}{\delta}}}{n_{h, s, a}}} + \sqrt{\frac{S}{n_{h, s, a}}} \leq 2\sqrt{\frac{S\log{\frac{4HSAn}{\delta}}}{n_{h, s, a}}} \leq 2\sqrt{\frac{S\log\frac{4HSAn}{\delta}}{n\Bar{d}_m}}
\end{equation}
where the final inequality reflects our exploration assumption on the logger. By the union bound, the above holds simultaneously for all $h, s, a$ with probability at least $1 - \delta/4$.

We bound $\|(\hat{V}^\pi_{h + 1} - V^\pi_{h + 1})(\cdot)\|_\infty$ using the Simulation Lemma again. We needn't be too careful here, because $(**)$ will decay with $\frac{1}{n} \ll \frac{1}{\sqrt{n}}$, and so be dominated by $(*)$.

Fix $h$ and $s$, and consider $\hat V^\pi_{h}(s) - V^\pi_{h}(s)$. Let $q^{(\pi, h, s)}_{t}(\cdot, \cdot)$ be the marginal visitation under policy $\pi$ and MDP, whose dynamics are defined by $P$, but which starts deterministically from  $(h, s)$ for $t \geq h$. 

\[\hat V^\pi_h(s) - V^\pi_h(s) = \sum_{t = h}^{H-1} \langle q^{(\pi, h, s)}_{t}(\cdot, \cdot), (\hat P_{t+1} - P_{t+1})^T\hat V^\pi_{t + 1} \rangle \leq \sum_{t = h}^{H - 1}\|q^{(\pi, h, s)}_{t}\|_1\|\hat{P}_{t + 1} - P_{t + 1}\|_\infty \|\hat{V}^\pi_{h + 1}\|_\infty\]
\[ \leq \sum_{t = h}^{H - 1} H \max_{s, a}\{\|(\hat{P}_{t + 1} - P_{t + 1})(\cdot | s, a)\|_1\} \leq 2H^2\sqrt{\frac{S\log\frac{4HSAn}{\delta}}{n\Bar{d}_m}},\]
where we have already provided for the control on $\max_{h, s, a}\{(\hat{P}_h(\cdot | s, a) - P_h(\cdot | s, a)\}$ in equation \eqref{1-norm control}.

Combining the results of this section, we achieve
\begin{equation}
    (**) \leq \frac{4H^2S\log\frac{HSAn}{\delta}}{n\Bar{d}_m}
\end{equation}

Combining the bounds on $(*)$ and $(**)$, we have that with probability at least $1- \delta/2$:
\[|\hat{v}^\pi\ - v^\pi| \leq \sum_{h = 1}^{H-1}\sum_{s, a}d^\pi_h(s, a)\left(\sqrt{\frac{2\Var[V^\pi_{h + 1}(s') | s' \sim P_{h + 1}(\cdot | s, a)]\log{\frac{4HSAn}{\delta}}}{n_{h, s, a}}} + \frac{4H}{3n\Bar{d}_m}\log{\frac{4HSAn}{\delta}}\right)\]
\[+ \frac{4H^3S\log{\frac{2HSAn}{\delta}}}{n\Bar{d}_m}.\]
Hence
\[|\hat{v}^\pi - v^\pi| \leq O(\sum_{h = 1}^{H - 1}\sum_{s, a} d^\pi_h(s, a)\sqrt{\frac{\Var[V^\pi_{h + 1}(s') | s' \sim P_{h + 1}(\cdot | s, a)]\log\frac{HSAn}{\delta}}{n_{h, s, a}}} + \frac{H^3S\log\frac{HSAn}{\delta}}{n\Bar{d}_m}).\]

An application of law of total variance \citep{azar17} leads to corollary \ref{pointwiseaope}. For proof of this cute fact, see, for example, Lemma 3.4 in \cite{yin20-0}.

\section{On the justification of considering $d_0$ and $r$ to be known}\label{wlog_sec}
In this section, we briefly justify our above assumption that $\hat{r} = r$ and $\hat{d}_0 = d_0$.

If, instead of using the true values $d_0$ and $r$, we use $\hat{d_0}$ and $\hat{r}$, the simulation lemma yields the following expansion of the error:
\[\hat v^\pi - v^\pi = \langle \hat d_0 - d_0, \hat V^\pi_0 \rangle \]\[+ \sum_{h = 1}^{H-1} \E_{s_h, a_h \sim d^\pi_h}[(\hat P_{h+1}(\cdot |s_h, a_h) - P_{h + 1}(\cdot | s_h, a_h))^T\hat V^\pi_{h + 1}(\cdot)] + \sum_{h = 1}^{H-1} \E_{s_h, a_h \sim d^\pi_h}[\hat r_h(s_h, a_h) - r_h(s_h, a_h)].\]

Applying Lemma \ref{d-dim conc} on the first term, we get a control on the form:
\begin{equation}\label{d0_est}|\langle \hat d_0 - d_0, \hat V^\pi_0 \rangle| \leq \|\hat{d}_0 - d_0\|_1\|\hat{V}^\pi_0\|_\infty \leq H(\sqrt{S/n} + \sqrt{\frac{\log\frac{1}{\delta}}{n}}).\end{equation}

Furthermore, for all $h, s, a$, we can control the reward-term by applying Hoeffding's Inequality, a union bound over $(h, s, a)$, and covering of $n_{h, s, a}$:
\begin{equation}\label{r_est}
    |\hat{r}_h(s, a) - r_h(s, a)| = |\frac{1}{n_{h, s, a}}\sum_{i: s^i_h = s, a^i_h = a} r^i_h - r_h(s, a)| \leq \sqrt{\frac{\log\frac{HSAn}{\delta}}{n_{h, s, a}}}.
\end{equation}

Both (\ref{d0_est}) and (\ref{r_est}) (summed over $H$) can be absorbed into the bounds of Corollaries \ref{iduniformaope} and \ref{idpointwiseaope}, at the cost of a constant factor.

\section{Proof of Lower Bound under Assumption \ref{expected_exploration} (Theorem \ref{lower_bound})}\label{lower_appendix}

\begin{figure}
\centering
  \includegraphics[width=.6\linewidth]{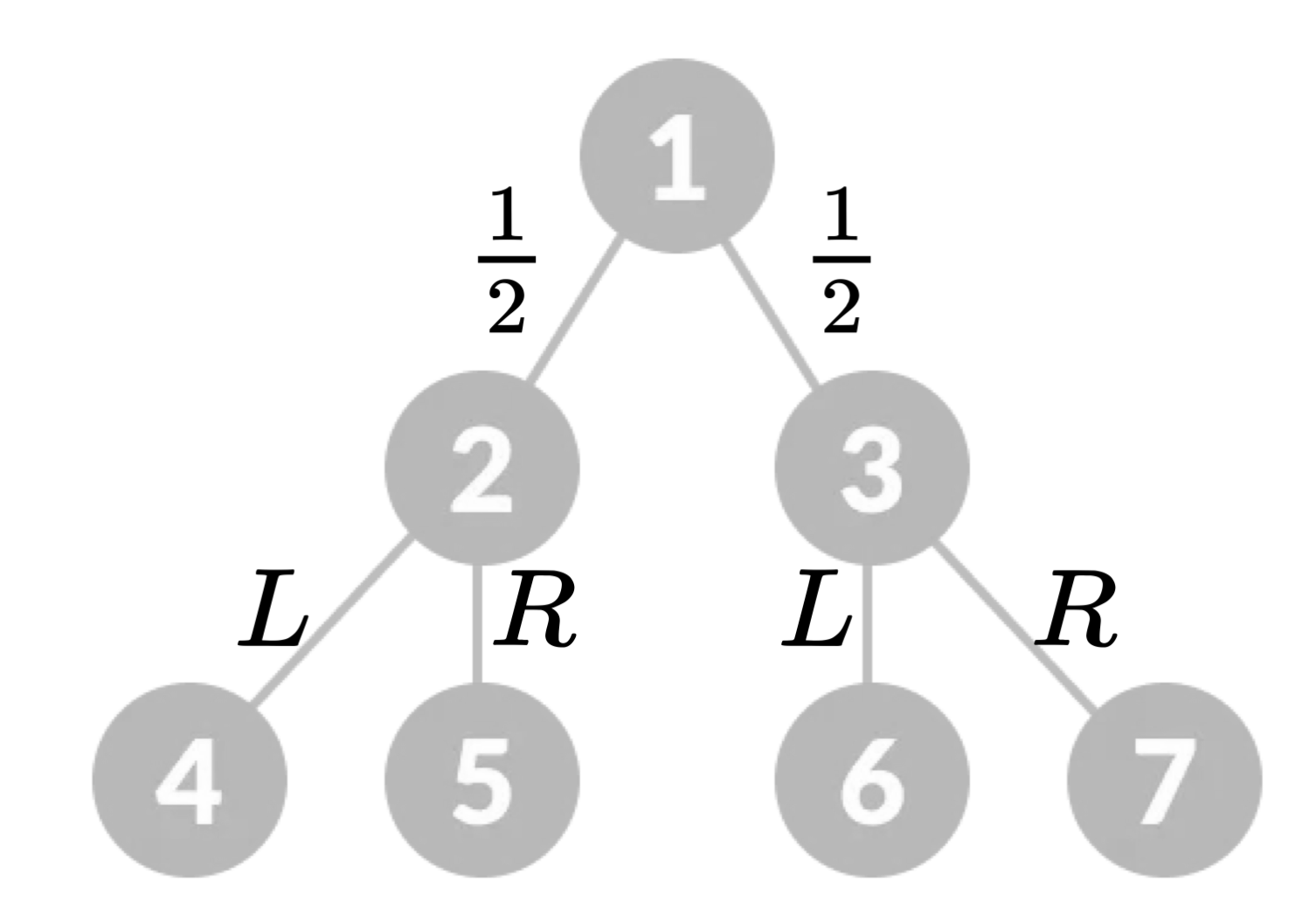}
\caption{Branching MDP construction used in the lower bound}
\label{lower_bound_dynamics}
\end{figure}

We consider the family of MDPs with dynamics as shown in Figure \ref{lower_bound_dynamics}. These MDPs have action space $\script A = \{L, R\}$, $H = 3$, and start deterministically at state $1$. From state $1$, the MDPs transition to state $2$ or state $3$ with equal probability regardless of action. From states $2$ and $3$, the MDPs transition left or the right deterministically according to the action chosen. 

\subsection*{Construction of Hard Instances}
We now specify an adaptive logger, $\script E$, for all MDPs with the above transition structure. $\script E$ begins with $\mu^1_h = L$ for all $h \in [H]$. If $\tau_1$ contains state $2$, $\script E$ chooses $\mu^j_h = L$ for all $h \in [H]$ and $j \in \{2, ... n\}$. Otherwise, $\mu^j_h = L$ for all $h \in [H]$ and $j$ even, and $\mu^j_h = R$ for all $h \in [H]$ and all $j$ odd. This construction satisfies Assumption \ref{expected_exploration} with parameter $\bar d_m > 0$.

\begin{figure}
\centering
  \includegraphics[width=\linewidth]{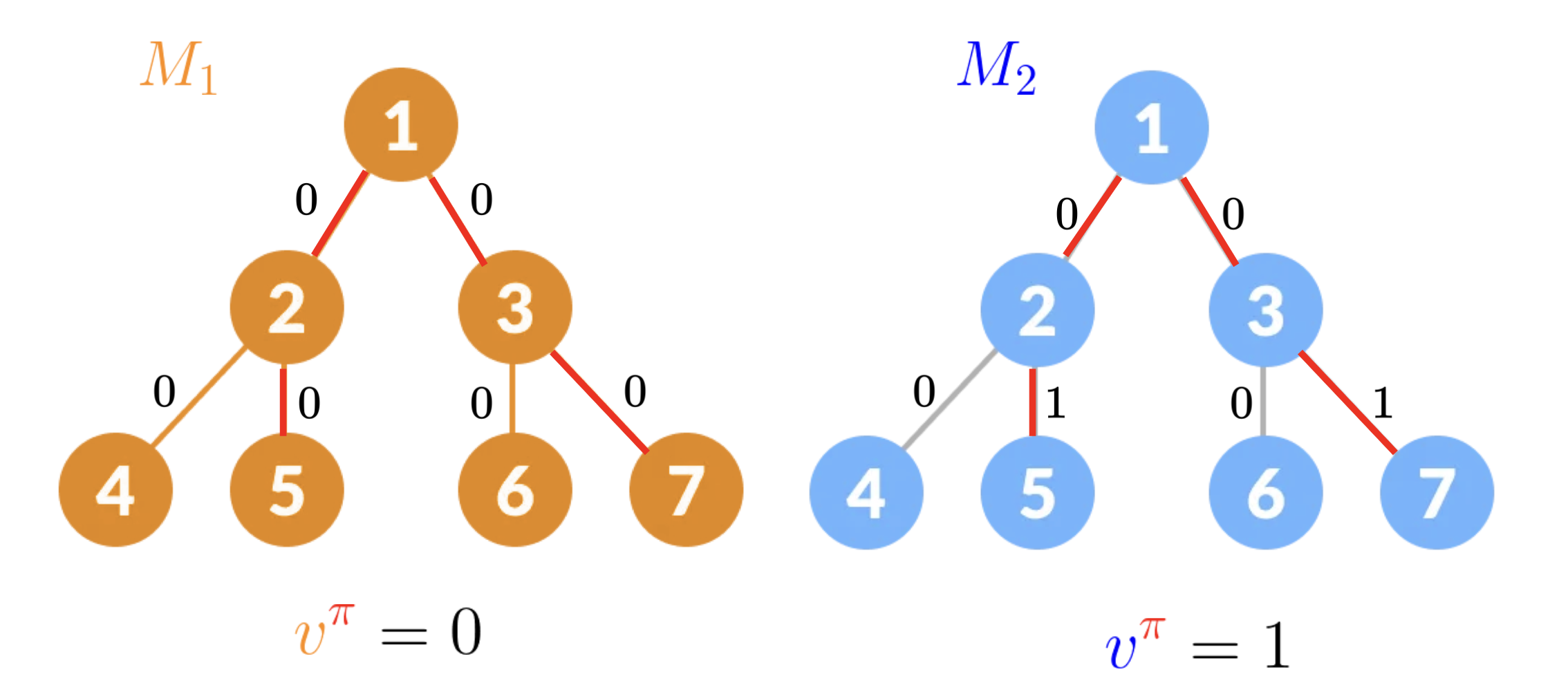}
\caption{Two ways of equipping the tree system with a reward function}
\label{lower_bound_reward}
\end{figure}

We now consider two MDPs with the same transition structure but with two different reward functions, as shown in Figure \ref{lower_bound_reward}, resulting in $M_1$ and $M_2$. $M_1$ deterministically administers reward $0$ for all transitions. $M_2$ deterministically administers reward $1$ for transitioning right at the second level, and reward $0$ otherwise.

The target policy $\pi$ always takes action $R$.

\subsection*{Packing}

We will show that no learning algorithm, $L$, can simultaneously do well on instances $I_1 = (M_1, \script E, \pi)$ and $I_2 = (M_2, \script E, \pi)$.

Define the event $B_1 := \{|L(\pi, \script D) - v^\pi| > \frac{1}{2}\}$ for $\script D \sim (M_1, \script E)$. Similarly, define $B_2 := \{|L(\pi, \script D) - v^\pi| > \frac{1}{2}\}$ for $\script D \sim (M_1, \script E)$. 

Assume that $\Pr[B_1] < \delta$ (otherwise, $L$ fails on $I_1$). Letting $\script E_L$ be the \emph{event} that $\script{E}$ goes from state $1$ to state $2$ in $\tau_1$, and $\script E_R = (\script E_L)^c$ We have: 
\[\delta > \Pr[B_1] = \Pr[B_1 | \script E_L]\Pr[\script E_L] + \Pr[B_1 | \script E_R]\Pr[\script E_R] = \]
\[\Pr[B_1 | \script E_L]\frac{1}{2} + \Pr[B_1 | \script E_R]\frac{1}{2} > \Pr[B_1 | \script E_L]\frac{1}{2}\]
where the probability is over $\script D \sim (M_1, \script E)$

Thus, conditioned on $\script E_L$, $L(\pi, \script D) \in B(0, \frac{1}{2})$ with probability at least $1 - 2\delta$. But note that conditioned on $\script E_L$, the data from $M_1$ and $M_2$ look the same. Thus, under data drawn from $(M_2, \script E)$, the same property holds. Thus, we have that $\Pr[B_2 | \script E_L] > 1 - 2\delta$. This gives:

\[\Pr[B_2] = \Pr[B_2 | \script E_L]\Pr[\script E_L] + \Pr[B_2 | \script E_R]\Pr[\script E_R] = \]
\[\Pr[B_2 | \script E_L]\frac{1}{2} + \Pr[B_2 | \script E_R]\frac{1}{2} > (1 - 2\delta)\frac{1}{2} = \frac{1}{2} - \delta\]
where the probability is over $\script D \sim (M_2, \script E)$

Choosing $\delta = \frac{1}{4}$ yields $\Pr[B_2] > \delta$.

This shows that $\Pr_{\script D \sim (M_1, \script E)}[B_1] < \delta \implies \Pr_{\script D \sim (M_2, \script E)}[B_2] > \delta$, and completes the proof of Theorem \ref{lower_bound}.

\begin{remark}
    We could prove the same result using trees of depth $1$ with deterministic transitions out of the first state, but would then need $\script E$ to choose stochastic logging policies. We find the above construction more pleasing, because the randomness in the data-collection process arises only due to the MDP.
\end{remark}

\section{Data-Collection Model for Empirical Simulations}\label{simulations appendix}
\begin{figure}[ht]
\vskip 0.2in
\begin{center} 
\centerline{\includegraphics[width=0.6\columnwidth]{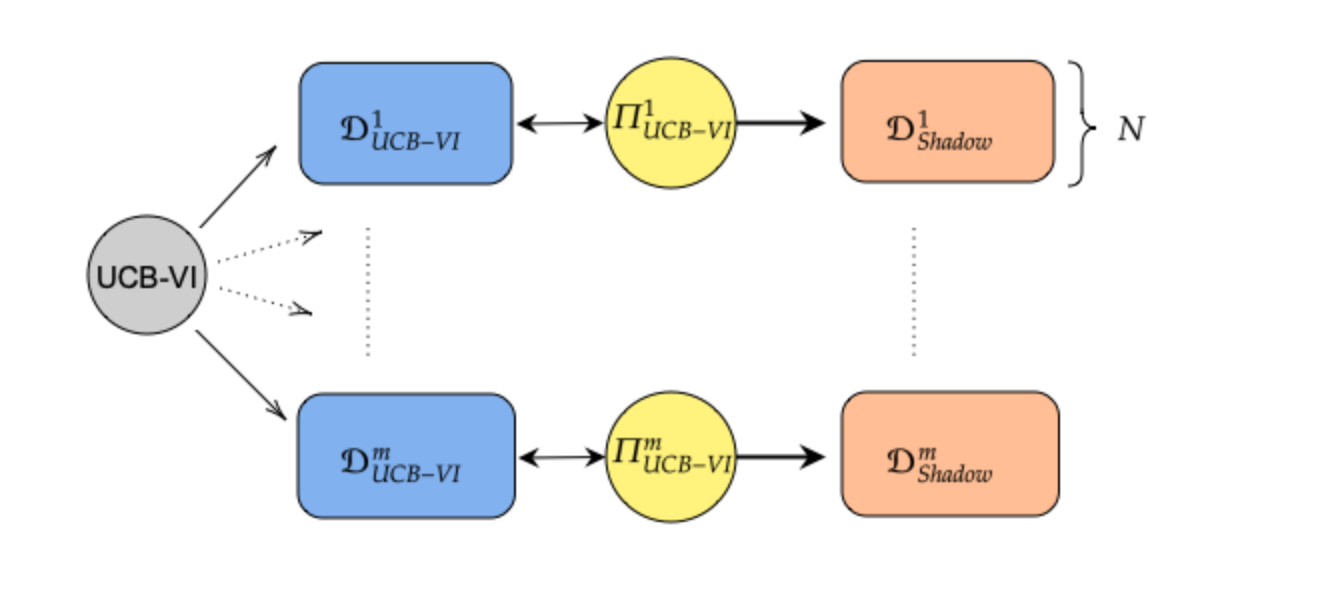}}
\caption{The data collection model for our experiments. We will instantiate $M = 10,000$ and $N = 1000$. Furthermore, note that each row starting from the grey to the orange specifies a process with outputs following some distribution.}
\label{subopt-plot}
\end{center}
\vskip -0.2in
\end{figure}

In order to generate copious amounts of data, we use a toy MDP with two states, two actions, and nonstationary dynamics and rewards over a time horizon of $H = 5$.. We choose as our logging algorithm UCB-VI \citep{azar17}. Running this algorithm for $N$ steps yields an adaptive dataset $\script{D}_{UCB-VI}$ generated according to logging policies $\Pi_{UCB-VI}$. We then collect a ``shadow'' dataset, $\script{D}_{Shadow}$, by rolling out $n$ independent trajectories $\{\tau'_i \sim \mu^i \ | \ \mu^i \in \Pi_{UCB-VI}\}$. We repeat this whole logging process many times, to generate several adaptive datasets and several shadow datasets, whose estimates we average. We then use this data to evaluate multiple policies.

As an optimism-based algorithm, UCB-VI is a good candidate for the testing of Question 1 above. Under optimism, it is likely that $\hat{v}^\pi$ is negatively biased  (especially for sub-optimal $\pi$), because in the event that the value of a state is substantially underestimated, we cease to gather data on that state. 

With regards to Question 2: As UCB-VI steers the data-collection procedure towards high-value states, it is conceivable that our estimator will benefit from the adaptivity for optimal $\pi$. UCB-VI can ``react'' to unwanted outcomes in trajectory $\tau_i$, where the Shadow dataset cannot.

\section{Multiple Independent Logging Policies}

In this section, we present some easy results relating to the Non-adaptive setting $\script D \sim (\mu^1, ... \mu^n)$ for $\{\mu^i\}$ independent. 

We warm up by generalizing \cite{yin20-0}'s bound on the MSE of $\hat{v}^\pi$ to NOPE. By following the proof of Theorem 1 in \cite{yin20-0}, and making some mild modifications, we recover a bound of the MSE of $\hat{v}^\pi$ in the Non-Adaptive OPE setting.

\begin{theorem}\label{nope}[MSE performance of $\hat v^\pi$ in NOPE setting]
Suppose $\script D$ is a dataset conforming to NOPE, and  $\hat v^\pi$ is formed using this dataset. 
Let $\bar d_m$ be as defined in \eqref{n-pol-dm}. Let $\tau_s = \max_{h, s, a}\frac{d^\pi_h(s, a)}{\frac{1}{n}\sum_id^{\mu^i}_h(s, a)}$. Let $\tau_a = \max_{h, s, a}\frac{\pi_h(a | s)}{\frac{1}{n}\sum_i \mu^i_h(a | s)}$. Then if $n > \frac{16\log{n}}{\bar d_m}$ and $n > \frac{4H\tau_a\tau_s}{\min_{h, s}\max{\{d^\pi_h(s), \frac{1}{n}\sum_i d^{\mu^i}_h(s)}\}}$, we have:
\begin{align*}
\mathrm{MSE}(\hat v^\pi) \leq& \widetilde{O}\left(\frac{1}{n}\sum_{h, s, a}\frac{d^\pi_h(s)^2\pi(a | s)^2}{\frac{1}{n}\sum_i{d^{\mu^i}_h(s, a)}}\psi_{h, s, a}\right) \\
&+ O(\tau_a^2\tau_sH^3/n^2\bar d_m),
\end{align*}
where $\psi_{h, s, a} := \Var[r^{(1)}_h + V_{h+1}^\pi(s^{(1)}_{h + 1}) | s^{(1)}_h,a^{(1)}_h = s,a]$.
\end{theorem}

As a corollary, consider a ``quasi-adaptive'' data collection process, where each logging-policy, $\mu^i$, is run twice, generating i.i.d. $\tau_i$ and $\tau_i'$. Suppose future logging-policies $\mu^{j > i}$ are chosen by some algorithm $\script E$ depending on $\tau_i$ but not $\tau_i'$. We can use the same $\hat v^\pi$-estimator to perform OPE with $\script D_{shadow} = \{\tau_i'\}$, as long as the estimator doesn't touch $\script D = \{\tau_i\}$. If we assume that average exploration is sufficient w.h.p. over the execution of $\script E$, we can bound the MSE in this quasi-adaptive case using Theorem \ref{nope}. Corollary \ref{shadow_corr} below tells us that, in those cases where we have control over the logging process, we can recover attractive, instance-dependent bounds by doubling the sample-complexity. 

\begin{corollary} \label{shadow_corr}
Let $\script E$ be the algorithm described in the paragraph above. Assume that with high probability ($\geq 1-\delta$), the policies $\mu^1, ..., \mu^n$ generated by $\script E$ satisfy $\frac{1}{n}\sum_id^{\mu^{i}}_h(s, a)  \geq \bar d_m$ for all $s\in \mathcal{S}$ and $a \in \mathcal{A}$, and for some $\bar d_m$. Then it holds that
\[\mathrm{MSE}(\hat v^\pi) \leq (1- \delta)(*) + H^2\delta,\] where $(*)$ is the bound on the MSE of the estimator in the non-adaptive case from Theorem \ref{nope}.
\end{corollary}

\subsection{Proof Outlines}
Theorem \ref{nope} is a straightforward generalization of the main result in \cite{yin20-0}. They use a multiplicative Chernoff bound to ensure that $n_{h, s, a} \geq nd_m$ for all $h, s, a$, with $d_m$ as defined in \eqref{sing-pol-dm} giving a lower-bound on success probability of $n_{h, s, a}$'s binomial distribution. If we instead use the Chernoff bound for the sum of independent Bernouli random variables with distinct success probabilities, we achieve $n_{h, s, a} \geq n\Bar{d}_m/2$ for all $h, s, a$ with high probability. By following the template for variance-analysis given by \cite{yin20-0}, the desired result can be recovered. Corollary \ref{shadow_corr} follows from Theorem \ref{nope} paired with the fact that the $\{\tau_i'\}$ are mutually independent conditioned on $\{\mu^i\}$ and the tower rule.
\end{document}